\documentclass[10pt,twocolumn,letterpaper]{article}

\usepackage{iccv}
\usepackage{times}
\usepackage{epsfig}
\usepackage{graphicx}
\usepackage{amsmath}
\usepackage{amssymb}
\usepackage{color}

\usepackage[pagebackref=true,breaklinks=true,letterpaper=true,colorlinks,bookmarks=false]{hyperref}

\iccvfinalcopy 

\def\iccvPaperID{340} 

\ificcvfinal\fi
\begin{document}

\title{Deep Temporal Appearance-Geometry Network\\for Facial Expression Recognition}

\author{Heechul Jung$^{\dagger}$~~~Sihaeng Lee$^{\dagger}$~~~Sunjeong Park$^{\dagger}$~~~Injae Lee$^{\ddagger}$~~~Chunghyun Ahn$^{\ddagger}$~~~Junmo Kim$^{\dagger}$\\
Korea Advanced Institute of Science and Technology$^{\dagger}$\\
Electronics and Telecommunications Research Institute$^{\ddagger}$\\
{\tt\small \{heechul, haeng, sunny0414, junmo.kim\}@kaist.ac.kr$^{\dagger}$, \{ninja, hyun\}@etri.re.kr$^{\ddagger}$}
}


\maketitle

\begin{abstract}
Temporal information can provide useful features for recognizing facial expressions.
However, to manually design useful features requires a lot of effort. In this paper, to reduce this effort, a deep learning technique which is regarded as a tool to automatically extract useful features from raw data, is adopted. Our deep network is based on two different models. The first deep network extracts temporal geometry features from temporal facial landmark points, while the other deep network extracts temporal appearance features from image sequences . These two models are combined in order to boost the performance of the facial expression recognition. Through several experiments, we showed that the two models cooperate with each other. As a result, we achieved superior performance to other state-of-the-art methods in CK+ and Oulu-CASIA databases. Furthermore, one of the main contributions of this paper is that our deep network catches the facial action points automatically.
\end{abstract}

\vspace{-4mm}
\section{Introduction}
\begin{figure}[t!]
\begin{center}
\includegraphics[width=6cm]{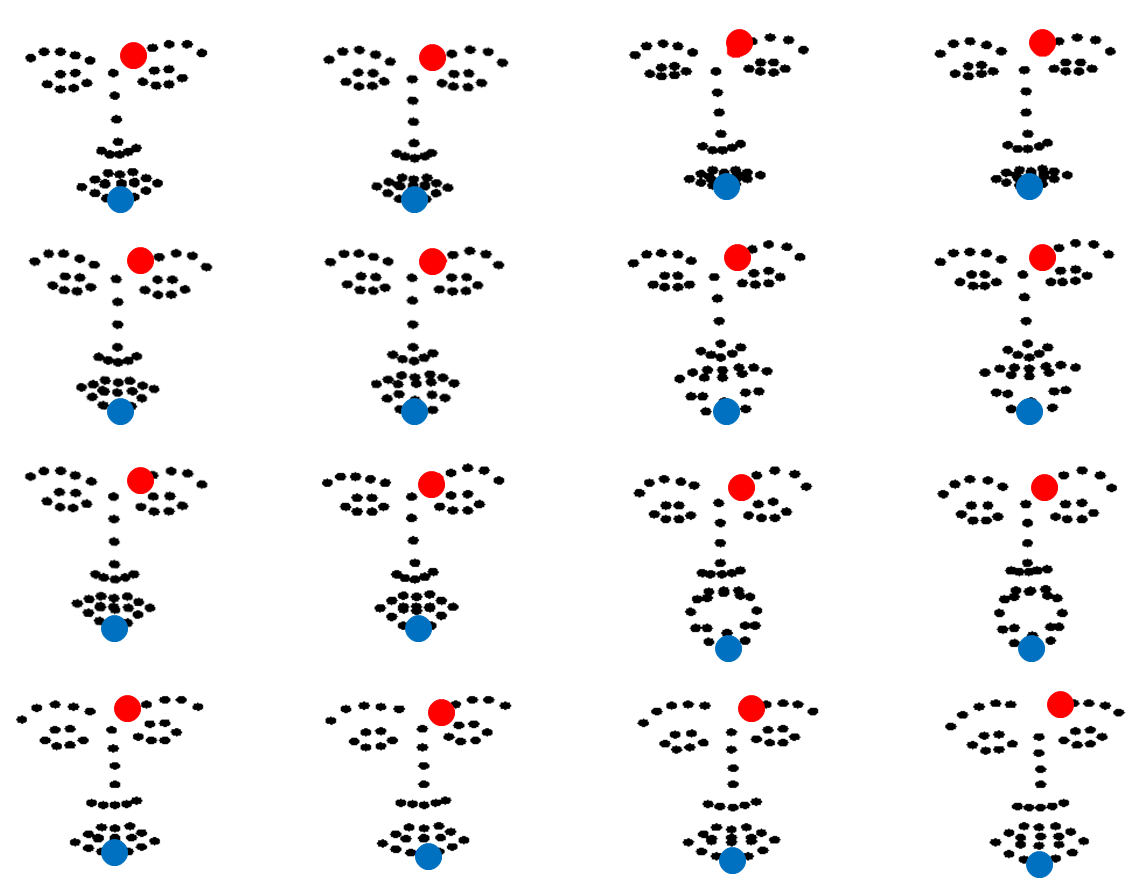}
\vspace{1mm}
\caption{\textbf{The Sequences of Facial Landmark Points.} Each row represents the sequences of the facial landmark points for each emotion. (Top to bottom: anger, happiness, surprise, and fear) The change of the vector from the red point to blue point over time can aid in facial expression recognition, but no one knows whether the representation is really helpful. In this paper, we use deep learning techniques to extract useful geometric representations rather than a heuristic feature.}
\label{fig:feature}
\end{center}
\vspace{-4mm}
\end{figure}

Recognizing an emotion from a facial image is a classic problem in the field of computer vision area, and many studies have been conducted. Recently, facial expression recognition research has been performed to increase the recognition performance by extracting useful temporal features from the image sequences \cite{sanin2013spatio,wang2013capturing,liu2014learning}. In general, such spatio-temporal feature extractors are manually designed, which is a difficult task. For example, Figure \ref{fig:feature} shows the facial landmark points of an image sequence with four emotions: anger, happiness, surprise, and fear. In order to classify the four emotions using the landmark points, one may use a temporal change of distance between blue and red points over time.
However, this representation can sometimes be ambiguous. This is just a guess, so we cannot assure that the information in the representation is really useful. 

Well-known deep learning algorithms, such as the deep neural network (DNN) and the convolutional neural network (CNN), have an ability to automatically extract useful representation from raw data (e.g., image data). However, there is a limit in applying it directly in facial expression recognition databases such as CK+ \cite{lucey2010extended}, MMI \cite{valstar2010induced} and Oulu-CASIA \cite{zhao2011facial}. The major reason is that the amount of data is very small, so a deep network that has many parameters can easily fall into overfitting when training a deep network. This is directly related to decreased accuracy.
Furthermore, if the training data is high dimensional, the overfitting problem is more crucial.

In this paper, we are interested in recognizing facial expressions using image sequence data with a deep network. In order to overcome the problem of a small amount of data, we construct two small deep networks that complement each other. One of the deep networks is trained using image sequences, while the other deep network learns the temporal trajectories of facial landmark points. In other words, the first network focuses more on appearance changes of facial expressions over time, while the second network is directly related to the motion of facial organs. We utilize CNN for training based on image sequences, and we use DNN for training based on temporal facial landmark points. Through this process, our two deep networks learn the emotional facial actions in different ways. Our main contributions in this paper are as follows:
\begin{itemize} \itemsep 1pt \parskip0pt \parsep0pt
\item Two deep network models are presented in order \textbf{to extract useful temporal representations from two kinds of sequential data}: image sequences and the trajectories of landmark points.
\item We observed that the two networks \textbf{automatically detect moving facial parts and action points}, respectively.
\item By integrating these two networks with different characteristics, \textbf{performance improvement} is achieved in terms of the recognition rates in both CK+ and Oulu-CASIA databases.
\end{itemize}

\begin{figure*}[t!]
\begin{center}
\includegraphics[width=16cm]{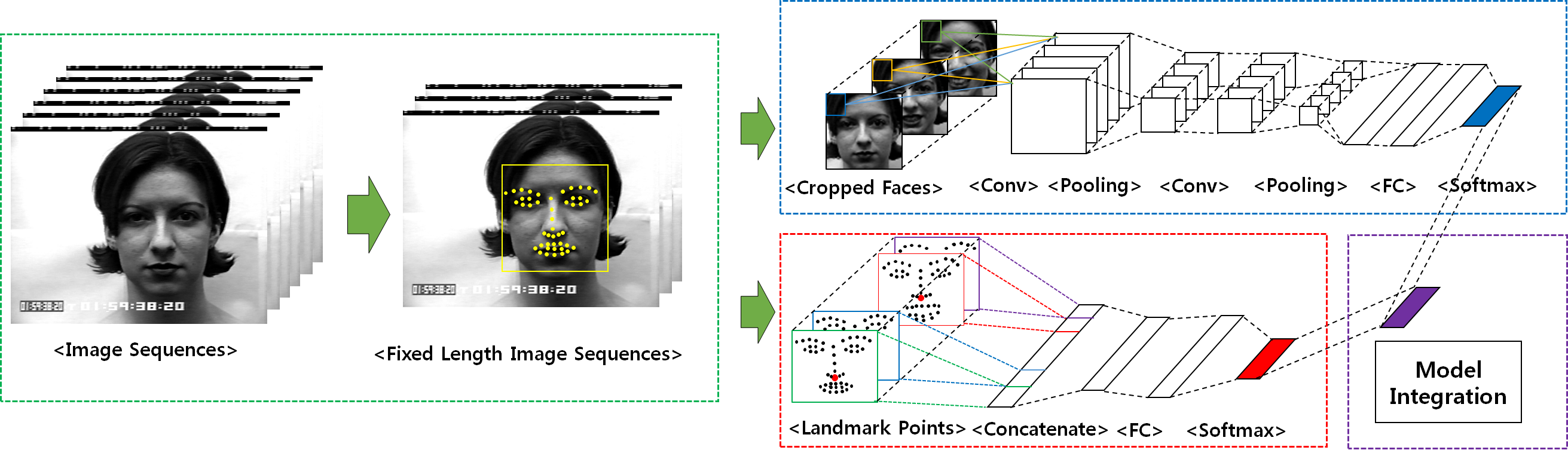}
\vspace{1mm}
\caption{\textbf{Overall Structure of Our Approach.} The green box with a dotted line represents the preprocessing step of our method. The blue and red boxes with dotted lines correspond to the two architectures of the deep networks. Our two deep networks receive an image sequence and facial landmark points as input, respectevely. Conv and FC refer to the convolutional and fully connected layers, respectively. Finally, the outputs of these networks are integrated using weighted summation, which is represented in the purple box.}
\label{fig:overview}
\end{center}
\vspace{-4mm}
\end{figure*}

Another advantage of our method compared to other state-of-the-art algorithms is that it is easy to implement using any deep learning codes, such as CudaConvnet2, theano and Caffe \cite{krizhevsky2014one, bergstra+al:2010-scipy, jia2014caffe}. Also, the codes of the preprocessing algorithms used in our method were already opened, and they can be easily obtained from their websites \cite{xiong2013supervised, opencv, zhou2011towards}. Furthermore, the number of parameters of our deep network is not many, and we utilize rectified linear unit (ReLU) \cite{glorot2011deep}, so the prediction can be performed fast. 
\section{Related Work}
\label{sec:related_work}
\subsection{Deep Learning-based Method}
Typically the CNN uses the single image, but CNN can also be used for temporal recognition problems such as action recognition \cite{sanin2013spatio}. In this 3D CNN method, the filters are shared along the time axis. Also, this method has been applied to facial expression recognition with deformable action part constraints, which is called 3D CNN-DAP \cite{liu2014deeply}. The 3D CNN-DAP method is basically based on 3D CNN and uses the strong spatial structural constraints of the dynamic action parts. It could receive a performance boost from using the hybrid method, but it falls short of the performance of other state-of-the-art methods.

\subsection{Hand-crafted Feature-based Method}
Many studies in this field have been conducted. Traditional local features such as HOG, SIFT, and LBP have been extended in order to be applicable to video, and these are called 3D HOG \cite{klaser2008spatio}, 3D SIFT \cite{scovanner20073} and LBP-TOP \cite{zhao2007dynamic}, respectively. Also, there was an attempt to improve accuracy through temporal modelings of each facial shape (TMS) \cite{jain2011facial}
. They used conditional random fields and shape-appearance features created by hand. 

Recently, spatio-temporal covariance decriptors with the Riemannian locality preserving projection approach were developed (Cov3D) \cite{sanin2013spatio}, and an interval temporal Bayesian network (ITBN) for capturing complex spatio-temporal relations among muscles was proposed \cite{wang2013capturing}. Also, most recently, expressionlet-based spatio-temporal manifold representation was developed (STM-ExpLet) \cite{liu2014learning}. The performance of this technique is best in CK+ and MMI databases.

\section{Motivation}
Human facial movement over time is very closely associated with emotions. For example, both ends of the lips go up when people make a happy expression.

Such movements of the face can be defined as a facial action coding system (FACS) \cite{facs}. In particular, there is an emotional facial action coding system (EFACS) to classify the movement of the face according to facial expression \cite{facs}. The majority of facial expression recognition databases follow this system.

Our approach is inspired by the EFACS. We are interested in finding meaningful temporal facial action features to improve the performance of facial expression recognition without using any prior information about EFACS. To achieve this, we adopt a deep learning technique to extract useful features automatically. However, it was difficult to catch the facial action information from the face using only CNN. Consequently, we also used temporal facial landmark points, which represent the movement of specific parts of face.

\section{Our Approach}
We utilize deep learning techniques in order to recognize facial expressions. Basically, two deep networks are combined: the deep temporal appearance network (DTAN) and the deep temporal geometry network (DTGN).
The DTAN, which is based on CNN, is used to extract the temporal appearance feature necessary for facial expression recognition. The DTGN, which is based on DNN, catches geometrically moving information from the facial landmark points. Finally, these two models are integrated in order to increase the expression recognition performance. This network is called the deep temporal appearence-geometry network (DTAGN). The flowchart for our overall process is shown in Figure \ref{fig:overview}.

\subsection{Preprocessing}
In general, the length of image sequences is variable, but the input dimension is usually fixed in a deep network. Consequently, the normalization along the time axis is required as input for the networks. We adopted the method in \cite{zhou2011towards}, which makes a image sequence into a fixed length.,

Then, the faces in the input image sequences are detected, cropped, and rescaled to 64$\times$64. From these detected faces, facial landmark points are extracted using the algorithm called IntraFace \cite{xiong2013supervised}. This algorithm provides accurate facial landmark points consisting of 48 landmark points, including two eyes, a nose, a mouth and two eyebrows.


\subsection{Deep Temporal Appearance Network}
In this paper, CNN is used for capturing temporal changes of appearance. Conventional CNN uses still images as input, and 3D CNN was presented recently for dealing with image sequences. As mentioned in Section \ref{sec:related_work}, the 3D CNN method shares the 3D filters along the time axis \cite{sanin2013spatio}.
However, we use the $n$-image sequences without weight sharing along the time axis. This means that each filter plays a different role depending on the time.
The activation value of the first layer is defined as follows:
\begin{equation}
f_{x,y,i} = \sigma(\sum_{t=1}^{T_a}\sum_{r=0}^{R}\sum_{s=0}^{S}I^{(t)}_{x+r,y+s}\cdot w_{r,s,i}^{(t)}+b_{i}),
\end{equation}
where $f_{x,y,i}$ is the activation value of position $(x,y)$ of the $i$-th feature map. $R$ and $S$ are the number of rows and columns of the filter, respectively. $T_a$ is the total frame number of the input grayscale image sequences. $I^{(t)}_{x+r,y+s}$ means that the value at the position $(x+r, y+s)$ of the input frame at time $t$. $w_{r,s,i}^{(t)}$ is the $i$-th filter value at $(r,s)$ for the $t$-th frame, and $b_i$ is the bias for the $i$-th filter.
$\sigma(\cdot)$ is an activation function, which is usually a non-linear function. Also, we utilize a ReLU, $\sigma(x)=max(0,x)$ as an activation function where $x$ is an input value \cite{glorot2011deep}.

The other layers are not different from the conventional CNN as follows: the output of the convolutional layer is rescaled to half-size in a pooling layer for efficient calculation. Using these activation values, a convolution operation and pooling are performed again. Finally, these output values are passed through the two fully connected layers and then classified using softmax.

For training our network, the stochastic gradient descent method is used for the optimization, and the dropout \cite{hinton2012improving} and weight decay methods are utilized for regularization.

Our network is not too deep and there are not many parameters to avoid overfitting, since the size of the facial expression recognition database is too small: there are only 205 sequences in the MMI database. Also, the first layer turns out to detect the temporal difference of the appearance over input image sequences as discussed in Section \ref{sec:visualization}.

\subsection{Deep Temporal Geometry Network}
DTGN receives the trajectories of facial landmark points as input. These trajectories can be considered as one-dimensional signals and defined as follows : 
\begin{equation}
X^{(t)}=
\begin{bmatrix}
x_1^{(t)} & y_1^{(t)} & x_2^{(t)} & y_2^{(t)} & \cdots & x_n^{(t)} & y_n^{(t)} \\
\end{bmatrix}^{\top},
\end{equation}
where $n$ is the total number of landmark points at frame $t$, and $X^{(t)}$ is a $2n$ dimensional vector at $t$. $x_k^{(t)}$ and $y_k^{(t)}$ are coordinates of the $k$-th facial landmark points at frame $t$.

These $xy$-coordinates are inappropriate for direct use as input in the deep network, because they are not normalized. For the normalization of the $xy$-coordinates, we first subtract the nose position of the face from each point (the position of the red point among the facial landmark points in the red box with the dotted line in Figure \ref{fig:overview}). Then, each coordinate is divided by each standard deviation of $xy$-coordinates in each frame as follows:
\begin{equation}
\bar{x_i}^{(t)}=\frac{x_i^{(t)}-x_o^{(t)}}{\sigma_x^{(t)}},
\end{equation}
where $\bar{x}_k^{(t)}$ is $x$-coordinate of the $k$-th facial landmark point at frame $t$, $x_o^{(t)}$ is $x$-coordinate of the nose landmark coordinate at frame $t$. $\sigma_x^{(t)}$ is standard deviation of $x$-coordinates at frame $t$. This process is also applied to the $y_i^{(t)}$.
Finally, these normalized points are concatenated along the time, and these points are used for the input to the DTGN.
\begin{equation}
\bar{X}=
\begin{bmatrix}
\bar{x}_1^{(1)} & \bar{y}_1^{(1)} & \cdots & \bar{x}_n^{(T_g)} & \bar{y}_n^{(T_g)} \\
\end{bmatrix}^{\top},
\end{equation}
where $\bar{X}$ is a $2nT_g$ dimensional input vector, and $\bar{x}_k^{(T_g)}$ and $\bar{y}_k^{(T_g)}$ are coordinates of $k$-th normalized landmark points at frame $T_g$.

The figure in the red box with a dotted line in Figure~\ref{fig:overview} illustrates the architecture of our DTGN model. Our network receives the concatenated landmark points $\bar{X}$ as input. Basically, we utilize two hidden layers, and the top layer is a softmax layer. Similar to the DTAN, this network is also trained by using the stochastic gradient descent method. The activation function for each hidden layer is ReLU. Furthermore, for regularization of the network, dropout \cite{hinton2012improving} and weight decay are used.

\subsection{Data Augmentation}
In order to better classify unobserved data, a number of training data are required. However, facial expression databases, such as CK+, Oulu-CASIA, and MMI, provide only sequences of hundreds. This makes a deep network easier to over-fit, because a deep network has many parameters. To prevent this problem, various data augmentation techniques are required.

First, whole image sequences are horizontally flipped. Then, each image is rotated according to each angle in $\{-15^\circ, -10^\circ, -5^\circ, 5^\circ, 10^\circ, 15^\circ \}$. This makes the model robust against the slight rotation changes of the input images. Finally, we obtain fourteen times more data: original images (1), flipped images (1), rotated images for each angle, and their flipped versions (12).

Similar to the augmentation of image sequences, the normalized facial landmark points are also horizontally flipped. Then, Gaussian noise is added to the raw landmark points.
\begin{equation}
\label{eq:da1}
\tilde{x}_{i}^{(t)}=\bar{x}_{i}^{(t)} + z_{i}^{(t)},
\end{equation}
where $z_{i}^{(t)} \sim N(0,\sigma_i^2)$ is additive noise with noise level $\sigma_i$ for the $x$-coordinate of the $i$-th landmark points at frame $t$. We set the value $\sigma_i$ to 0.01. Also, we contaminated $y$-coordinate with noise in the same way. This method prevents slight pose changes.

To prepare for rotation changes, we construct rotated the data as follows: 
\begin{equation}
\label{eq:da2}
\begin{bmatrix}
\tilde{x}_i^{(t)} & \tilde{y}_i^{(t)}
\end{bmatrix}^{\top}
= R^{(t)}
\begin{bmatrix}
\bar{x}_i^{(t)} & \bar{y}_i^{(t)}
\end{bmatrix}^{\top},
\end{equation}
for $i=1,\ldots,n$ where $\tilde{x}_j^{(t)}$ and $\tilde{y}_j^{(t)}$ are $j$-th rotated $xy$-coordinates at time $t$, and $R^{(t)}$ is a 2 $\times$ 2 rotation matrix for the $xy$-coordinates at time $t$, which has an angle $\theta^{(t)}$. The value of $\theta^{(t)}$ is drawn from a uniform distribution where $\theta^{(t)} \sim Unif[\beta,\gamma]$.
We set the values $\beta$ and $\gamma$ to $-\pi / 10$ and $\pi / 10$, respectively.


 We performed the first data augmentation methods in equation \ref{eq:da1} three times, and the second data augmentation in equation \ref{eq:da2} was also conducted three times. Consequently, we obtained six times more facial landmark points. As a result, we augmented the training data fourteen times: original coordinates (1), flipped coordinates (1) and two augmentation methods for each (12).

\begin{figure}[t!]
\begin{center}
\includegraphics[width=3cm]{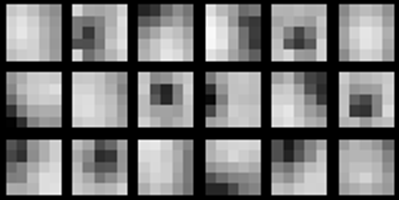}
\vspace{1mm}
\caption{\textbf{Filters Learned by Single Frame-based CNN.} The input image size was 64 $\times$ 64, and the filter size was 5 $\times$ 5. 18 filters were selected for visualization from 64 learned filters in the first convolutional layer. The black and white colors represent the negative and positive values, respectively. There were several directional edge and blob detection filters.}
\label{fig:filter_org}
\end{center}
\vspace{-4mm}
\end{figure}

\begin{figure}[t!]
\begin{center}
\includegraphics[width=\linewidth]{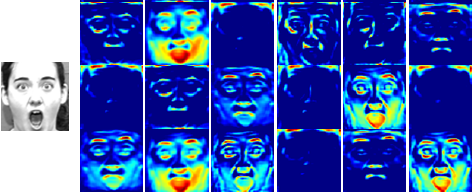}
\caption{\textbf{Feature Maps Corresponding to Figure \ref{fig:filter_org}.}  The left image represents the input image, and the right image shows the feature maps extracted by each filter in Figure \ref{fig:filter_org}. The emotion label for the input image was surprise. The blue and red values represent the low and high response values, respectively. The edges of the input image are detected in most of the filter.}
\label{fig:feature_org}
\end{center}
\vspace{-4mm}
\end{figure}

\subsection{Model Integration}
\label{sec:mi}
The outputs from the top layers of the two networks were integrated using equation \ref{eq:avg}.
\begin{equation}
\label{eq:avg}
o_i = \alpha p_i + (1-\alpha) q_i,~0 \leq \alpha \leq 1,
\end{equation}
for $i=1,\ldots,c$ where $c$ is the total number of emotion class, $p_i, q_i$ are each output of DTAN and DTGN, and $o_i$ is the final score. Finally, the index with the maximum value is the final prediction. This parameter $\alpha$ depends on the performance of each network. If the performances of the two networks are similar to each other, the value of $\alpha$ is 0.5 (e.g., the experiments on the CK+ and Oulu-CASIA).


\begin{figure}[t!]
\begin{center}
\includegraphics[width=\linewidth]{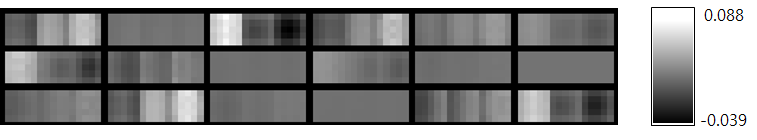}
\caption{\textbf{Filters Learned by DTAN.} The number of input frames was three in this figure, so there are three filters corresponding to each frame. The three filters in each bold black box generate one feature map. As with Figure \ref{fig:filter_org}, 18 filters were selected from 64 learned filters. In this figure, we guess that our network detects differences between frames.}
\label{fig:filter_cnn}
\end{center}
\vspace{-4mm}
\end{figure}

\begin{figure}[t!]
\begin{center}
\includegraphics[width=\linewidth]{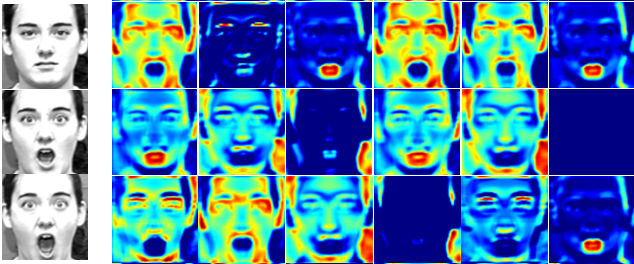}
\caption{\textbf{Feature Maps Corresponding to Figure \ref{fig:filter_cnn}.} The gray images on the left side form the image sequence used as input, and the images on the right side are the feature maps corresponding to each filter in Figure \ref{fig:filter_cnn}. Blue and red represent the low and high response values. The emotion label for the input image sequence was surprise. We observed that our network responded to moving parts for expressing emotion.}
\label{fig:feature_cnn}
\end{center}
\vspace{-4mm}
\end{figure}

\begin{figure*}[t!]
\begin{center}
\includegraphics[width=15cm]{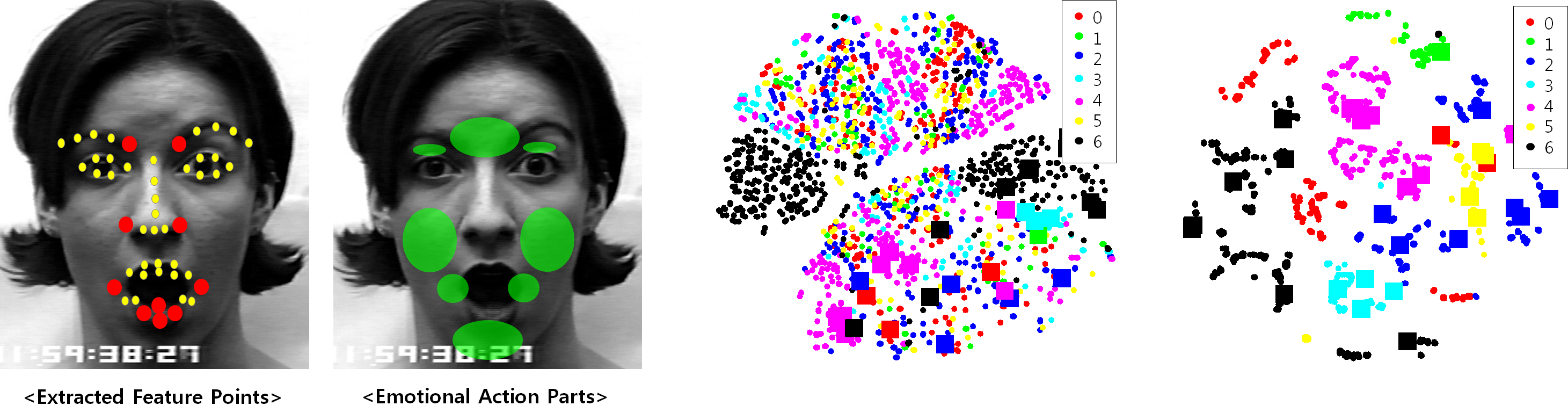}
\end{center}
\vspace{-4mm}
~~~~~~~~~~~~~~~~~~~~~~~~~~~~~~~~~~~~~~~~~~~~~~~~(a)~~~~~~~~~~~~~~~~~~~~~~~~~~~~~~~~~~~~~~~~~~~~~~~~~~~~~~(b)~~~~~~~~~~~~~~~~~~~~~~~~~~~~~~~~~~~~~~~~~~~~~~~~~(c)
\vspace{-2mm}
\begin{center}
\vspace{1mm}
\caption{\textbf{Visualization of Representation Extracted by DTGN} (a) The important top-10 positions detected by our network (red points in the left figure). In order to visualize these ten points, we calculated the average of the absolute values of weights in the first FC layer connected to each landmark point. Then, these values were sorted by descending order, and top 10 points with highest value were selected. The action parts defined by EFACS \cite{facs} are shown in the right figure (green colored area). (b) Visualization of original input data in CK+ database, using t-SNE \cite{van2008visualizing}. The number of data was 4149: (327-33)$\times$14 augmented training data and 33 test data. The small dots and large squares represent training and test data, respectively. The numbers in the legend correspond to each label of the CK+ database: 0-anger, 1-contempt, 2-disgust, 3-fear, 4-happiness, 5-sadness, and 6-surprise. (c) Visualization of the outputs in the second hidden layer. The data points were automatically grouped by DTGN.}
\label{fig:tsne}
\end{center}
\vspace{-4mm}
\end{figure*}

\section{What Will Deep Networks Learn?}
\label{sec:visualization}
\subsection{Visualization of DTAN}
To find out what our DTAN learned, the learned filters were visualized. Figure \ref{fig:filter_org} demonstrates the filters learned by single frame-based CNN in the first convolutional layers using the CK+ database. The filters were similar to the edge or blob detectors. Corresponding responses to each filter are provided in Figure \ref{fig:feature_org}. The edge components according to the direction were detected by each filter.

In our DTAN, which is a multiple frame-based CNN, the learned filters are shown in Figure \ref{fig:filter_cnn}. Unlike the filters of single frame-based CNN, the filters were not edge or blob detectors. To exaggerate a little, these were just combinations of black, gray, and white filters. Figure \ref{fig:feature_cnn} shows the meaning of these filters. High response values ​​were usually shown in parts with big differences between input frames. In other words, we know that the first convolutional layer of our DTAN detects facial movements arising from the expression of emotion.

\subsection{Visualization of DTGN}
The left side of Figure \ref{fig:tsne} (a) shows the significant facial landmark points for facial expression recognition. These positions were automatically found by DTGN. The extracted positions were very similar those of EFACS on the right side in Figure \ref{fig:tsne} (a). To explain it further, the two extracted points on the nose become wider when people make a
happy expression because both cheeks are pulled up. Also, our network did not catch the eyelid, because the database includes the eye blinking action.

In order to figure out the characteristics of the features extracted from the top layer, we also visualized the augmented input feature vectors using t-SNE, which is a useful tool for visualization of high dimensional data \cite{van2008visualizing}. The input data were spread randomly in Figure \ref{fig:tsne} (b), but the features extracted from the second hidden layer were separated according to their label, as shown in Figure \ref{fig:tsne} (c).

\section{Experiments}
For assessing the performance of our method, we used three databases: the CK+, Oulu-CASIA, and MMI databases. The number of image sequences in each database are listed according to each emotion in Table~\ref{tb:db}.
In the experiments, the algorithms which were not mentioned in Section  \ref{sec:related_work} were used such as manifold based sparse representation (MSR) \cite{ptucha2011manifold}, AdaLBP \cite{zhao2011facial}, Atlases \cite{guo2012dynamic}, and common and specific active patches (CSPL) \cite{zhong2012learning}.

\begin{table}[b!]
\begin{center}
\begin{tabular}{c||ccccccc|c}
\hline
& An & Co & Di & Fe & Ha & Sa & Su & All\\ 
\hline
\hline
CK+& 45 & 18 & 59 & 25 & 69 & 28 & 83 & 327\\
Oulu&  80 & - & 80 & 80 & 80 & 80 & 80 & 480\\
MMI& 32 & - & 31 & 28 & 42 & 32 & 40 & 205\\
\hline
\end{tabular}
\end{center}
\caption{\textbf{The Number of Image Sequences for Each Emotion}: anger (An), contempt (Co), disgust (Di), fear (Fe), happiness (Ha), sadness (Sa), and surprise (Su).}
\label{tb:db}
\end{table}

\begin{table*}[t!]
\begin{center}
\begin{tabular}{c||cccccc|ccc}
\hline
& HOG 3D & MSR & TMS & Cov3D & STM-ExpLet & 3DCNN-DAP & \bf{DTAN} & \bf{DTGN} & \bf{DTAGN}\\
\hline\hline
Accuracy&91.44 & 91.4 & 91.89 & 92.3 & \color{blue}{\textbf{94.19}} & 92.4 & 91.44 & 92.35 & \color{red}{\textbf{96.94}}\\
\hline
\end{tabular}
\end{center}
\caption{\textbf{Overall Accuracy in the CK+ Database.} The red and blue colors represent the first and second most accurate, respectively.}
\label{tb:ck_result}
\end{table*}

\subsection{Implementation}
First, we normalized the original image sequences to twelve frames using the code downloded from \cite{zhou2011towards} for input. In the case of MMI, we normalized twenty four frames, because the total number of frames in the original image sequences was more than in the other databases. Then, we selected twelve frames on the front side. In order to detect a face in the normalized image sequences, we used the OpenCV Haar-like detector \cite{opencv}, and the detected faces were used for the initial position of the IntraFace \cite{xiong2013supervised}. For DTAN, we selected three face images (1st, 7th, and 12th frames). The architecture of our deep network was implemented by CudaConvnet2 \cite{krizhevsky2014one}. 

\begin{table}[b!]
\begin{center}
\begin{tabular}{c||ccccccc}
\hline
& An & Co & Di & Fe & Ha & Sa & Su\\ 
\hline
\hline
An&   \textbf{100} &      0 &    0 &  0 &  0 &  0 & 0\\
Co&  0 &  \textbf{94.44} &    0 &  0 &   0 &  5.56 & 0\\
Di&     0 &      0 &  \textbf{100} &   0 &   0 &  0 & 0\\
Fe&     0 &      0 &    0 &  \textbf{84} &  8 & 0 & 8\\
Ha&     0 &      0 &    0 &   0 &  \textbf{100} &        0 & 0\\
Sa& 10.71 &      0 &    0 &        3.57 &        0 &  \textbf{85.71} & 0\\
Su&     0 &    1.2 &    0 &        0 &        0 &  0 & \textbf{98.8}\\
\hline
\end{tabular}
\end{center}
\caption{\textbf{Confusion Matrix for the CK+ Database.}  The labels in the leftmost column and on the top represent the ground truth and prediction results, respectively.}
\label{tb:ck_cm}
\vspace{-4mm}
\end{table}

\subsection{CK+}
\noindent\textbf{Description of the Database.} CK+ is a representative database for facial expression recognition. This database is composed of 327 image sequences with seven emotion labels: anger, contempt, disgust, fear, happiness, sadness, and surprise. There are 118 subjects, and these subjects are divided into ten groups by ID ascending order. Nine subsets were used for training our networks, and the remaining subset was used for validation. This process is the same as the 10-fold validation protocol in \cite{liu2014learning}. In this database, each sequence starts with a neutral emotion and ends with a peak of the emotion.
\vspace{1.6mm}

\noindent\textbf{Details of the Architecture.}
The architecture of DTGN is D1176-FC100-FC600-S7. D1176 is a 1176 dimensional input vector, and FC100 refers to a fully connected layer with 100 nodes. Also, S7 is the softmax layer with seven outputs.
The rates for dropout were set to 0.1, 0.5, 0.5 for input and two hidden layers, respectively.

Our DTAN model for CK+ is I64-C(5,64)-L5-P2-C(5,64)-L3-P2-FC500-FC500-S7, where I64 means 64 $\times$ 64 input image sequences, and C(5,64) is a convolutional layer with 64 $5 \times 5$ filters. L5 is a local contrast normalization layer with a window size of $5 \times 5$. P2 means a $2 \times 2$ max pooling layer. The stride of each layer was the same as 1 with the exception of the pooling layer. The value of the stride for each pooling layer was set to 2. The $\alpha$ mentioned in Section \ref{sec:mi} was set to 0.5.
The dropout rates for DTAN were set to 0.1, 0.5, 0.5 for input and two fully connected layers, respectively.
\vspace{1.6mm}

\begin{table*}[t!]
\begin{center}
\begin{tabular}{c||cccccc|ccc}
\hline
& 3D SIFT & LBP-TOP & HOG 3D & AdaLBP & Atlases & STM-ExpLet & \bf{DTAN} & \bf{DTGN} & \bf{DTAGN}\\
\hline\hline
Accuracy& 55.83 & 68.13 & 70.63 & 73.54 & \color{blue}{\textbf{75.52}}  & 74.59 & 74.38 & 74.17 & \color{red}{\textbf{80.62}}\\
\hline
\end{tabular}
\end{center}
\caption{\textbf{Overall Accuracy in the Oulu-CASIA Database.} The red and blue colors represent the first and second most accurate, respectively.}
\label{tb:accuracy_oulu}
\end{table*}
\noindent\textbf{Results.} The total accuracy of 10-fold cross validation is shown in Table \ref{tb:ck_result}. The performances of DTAN and DTGN are lower than other algorithms, but the performance of the integrated network is better than other state-of-the-art algorithms.

\begin{figure}[t!]
\begin{center}
\includegraphics[width=\linewidth]{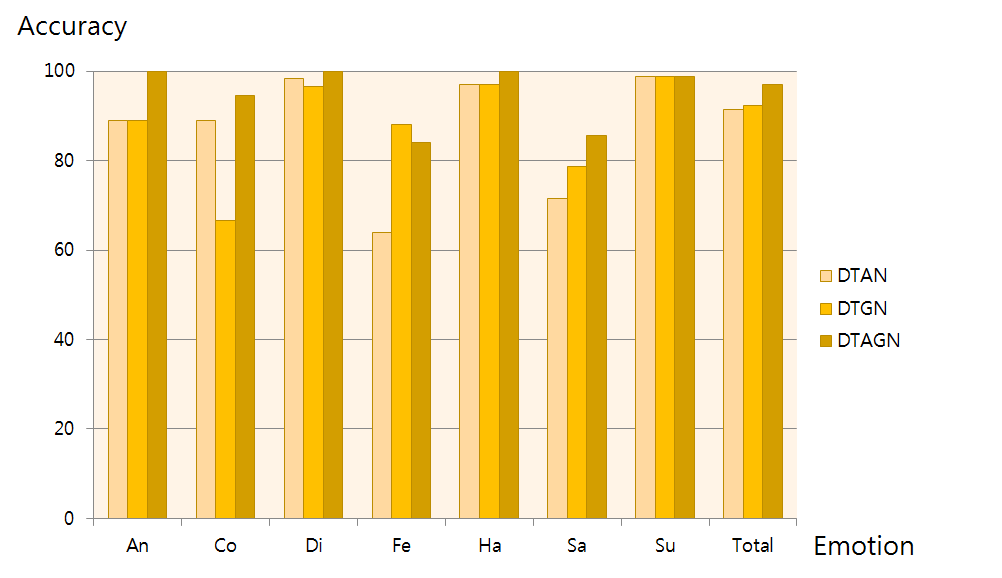}
\caption{\textbf{Comparison of Accuracy in the CK+} according to each emotion among three networks.}
\label{fig:ck_compare}
\end{center}
\vspace{-4mm}
\end{figure}

The two networks were complementary, and this is shown in Figure \ref{fig:ck_compare}. The DTAN had a good performance with respect to contempt, whereas it had lower accuracy with fear.
On the other hand, the geometry-based model was strong with fear.

 Table \ref{tb:ck_cm} shows the confusion matrix for CK+. Our algorithm has performed well in recognizing anger, disgust, happiness, and surprise. For the other emotions, our method also performed well.

\subsection{Oulu-CASIA}
\noindent\textbf{Description of the Database.} For further experiments, we used Oulu-CASIA, which includes 480 image sequences taken under the normal illumination condition. Each image sequence has one of six emotion labels: anger, disgust, fear, happiness, sadness, or surprise. There were 80 subjects, and the 10-fold validation was performed in the same way as in the case of CK+. Similar to the CK+ database, each sequence begins with a neutral facial expression and ends with the facial expression of each emotion.
\vspace{1.6mm}

\noindent\textbf{Details of the Architecture.}
For Oulu-CASIA, the architecture of DTGN was D1176-FC100-FC600-S6.
The DTAN model was also the same as the DTAN model of CK+ except the number of nodes in the top layer, because there are six labels in Oulu-CASIA. The value of $\alpha$ was set to 0.5. The dropout rates for DTAN and DTGN were the same as the rates in CK+.
\vspace{1.6mm}

\noindent\textbf{Results.} The accuracy of our algorithm was superior to the other state-of-the-art algorithms, as shown in Table \ref{tb:accuracy_oulu}. The best performance from among the existing methods was 75.52\%, which was achieved by Atlases, and this record had not been broken for three years. However, we have significantly improved the accuracy by about 5\%.

\begin{figure}[t!]
\begin{center}
\includegraphics[width=\linewidth]{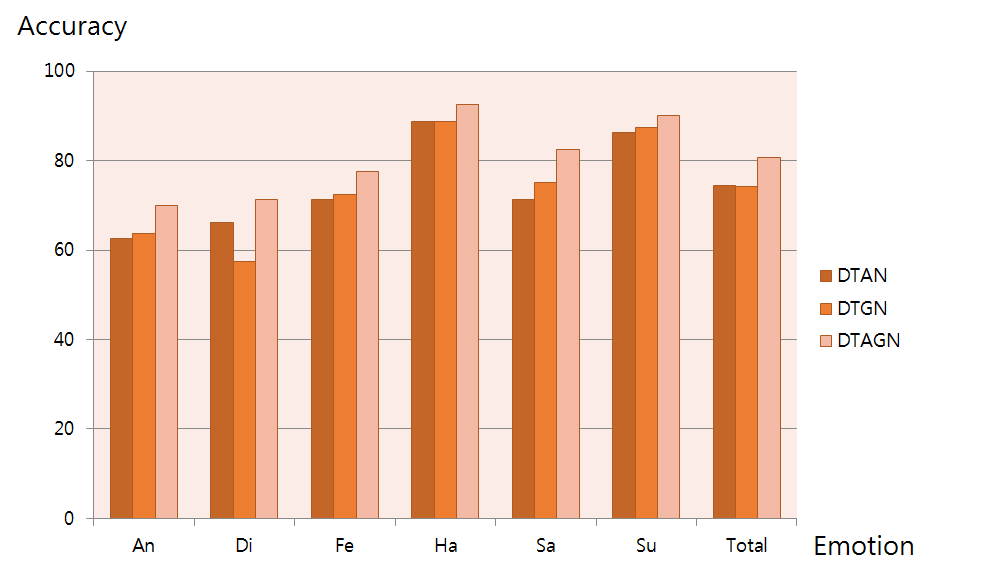}
\caption{\textbf{Comparison of Accuracy in the Oulu-CASIA} according to each emotion among three networks.}
\label{fig:oulu_compare}
\end{center}
\vspace{-4mm}
\end{figure}

In Figure \ref{fig:oulu_compare}, the performance of two networks was compared. Similar to the case of CK+, we observed that the two networks were complementary to each other. In particular, the performance of the DTGN in the case of disgust was lower than the DTAN, but the combined model produced good results. 

Table \ref{tb:oulu_cm} shows the confusion matrix for our algorithm. The performance in the cases of happiness, sadness, and surprise was good, but the performance for anger, disgust, and fear was relatively poor. In particular, anger and disgust were confused in our algorithm.

\begin{table}[b!]
\begin{center}
\begin{tabular}{c||cccccc}
\hline
& An & Di & Fe & Ha & Sa & Su\\ 
\hline\hline
An&      \textbf{70} &     15 &   3.75 &    1.25 &    10 &        0\\
Di&      20 &  \textbf{71.25} &      7 &       0 &     3.75 &        0\\
Fe&     2.5 &    2.5 &   \textbf{77.5} &    6.25 &     2.5 &   8.75\\
Ha&       0 &      0 &      5 &    \textbf{92.5} &     2.5 &        0\\
Sa&   11.25 &    2.5 &   3.75 &       0 &  \textbf{82.5} &   0\\
Su&      0  &      0 &     10 &       0 &        0 &  \textbf{90}\\
\hline
         
\end{tabular}
\end{center}
\caption{\textbf{Confusion Matrix for the Oulu-CASIA Database.} The labels in the leftmost column and on the top represent the ground truth and prediction results, respectively.}
\label{tb:oulu_cm}
\end{table}

\begin{table*}[t!]
\begin{center}
\begin{tabular}{c||cccccc|ccc}
\hline
& HOG 3D & 3D SIFT & ITBN & CSPL & STM-ExpLet & 3DCNN-DAP & \bf{DTAN} & \bf{DTGN} & \bf{DTAGN}\\
\hline\hline
Accuracy&60.89 & 64.39 & 59.7 & (73.53) & \color{red}{\textbf{75.12}} & 63.4 & 58.05 & 56.1 & \color{blue}{\textbf{66.33}}\\
\hline
\end{tabular}
\end{center}
\caption{\textbf{Overall Accuracy in the MMI Database.} The red and blue colors represent the first and second most accurate, respectively. The CSPL used additional ground truth information, so it was excluded from the ranking.}
\label{tb:accuracy_mmi}
\end{table*}

\subsection{MMI}
\noindent\textbf{Description of the Database.} MMI consists of 205 image sequences with frontal faces and includes only 30 subjects. Similar to the Oulu-CASIA database, there are six kind of emotion labels. This database was also divided into ten groups for person independent 10-fold cross validatiaon.

This database is different to the other databases, each sequence begins with a neutral facial expression, and has the facial expression of each emotion in the middle of the seqence. This ends with the neutral facial expression. The peak frame was not provided as a prior information.
\vspace{1.6mm}

\noindent\textbf{Details of the Architecture.}
We used the DTGN model of D1176-FC100-FC70-FC6 for the MMI database. The number of subjects and image sequences was very small, so we decreased the number of nodes significantly.

Our DTAN model was designed as I64-C(7,64)-L5-P2-C(5,32)-L3-P2-C(3,32)-L3-P2-FC300-FC300-6.
Differing from the other two databases, this database had many pose changes. Also, there were a variety of environments, such as like lighting changes. Consequently, we normalized the input faces using the eye coordinates, and local contrast normalization was used for input image sequences.  The dropout rates for DTAN and DTGN were the same as the rates in CK+.

The value of $\alpha$ was different than in the other experiments. We set the value as 0.42, and this value was manually determined.
\vspace{1.6mm}

\begin{figure}[t!]
\begin{center}
\includegraphics[width=\linewidth]{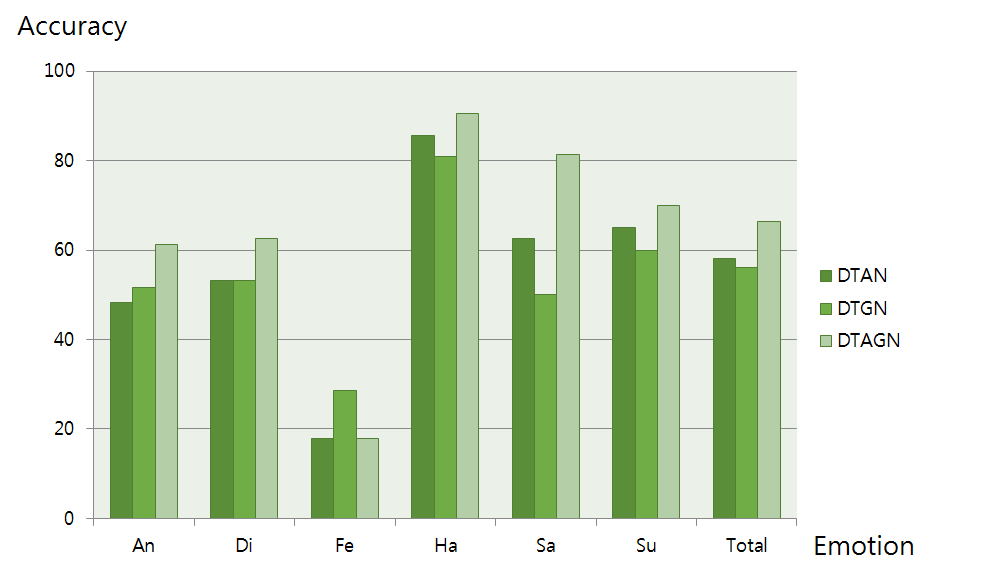}
\caption{\textbf{Comparison of Accuracy in the MMI} according to each emotion among three networks.}
\label{fig:mmi_compare}
\end{center}
\vspace{-4mm}
\end{figure}

\begin{table}[b!]
\begin{center}
\begin{tabular}{c||cccccc}
\hline
& An & Di & Fe & Ha & Sa & Su\\ 
\hline
\hline
An &  \textbf{61.29} &   22.6 &      0 &      0 &  12.9  &    3.23\\
Di &  21.88 &   \textbf{62.5} &      0 &      0 &   12.5 &    3.13\\
Fe &   3.57 &   3.57 &  \textbf{17.86} &   7.14 &   10.71 &   57.14\\
He &     0  &      0 &   4.76 &  \textbf{90.48} &      0  &    4.76\\
Sa &   3.13 &   3.13 &   6.25 &      0 &   \textbf{81.25} &    6.25\\
Su &    2.5 &      0 &     25 &    2.5 &        0 &   \textbf{70}\\
\hline
\end{tabular}
\end{center}
\caption{\textbf{Confusion Matrix for the MMI Database.} The labels in the leftmost column and on the top represent the ground truth and prediction results, respectively.}
\label{tb:mmi_cm}
\end{table}

\noindent\textbf{Results.}
In Table \ref{tb:accuracy_mmi}, our algorithm was good as the second. The CSPL algorithm was excluded for the ranking, because the CSPL used the peak frame number, which is an additional type of ground truth information.

\begin{figure}[t!]
\begin{center}
\includegraphics[width=\linewidth]{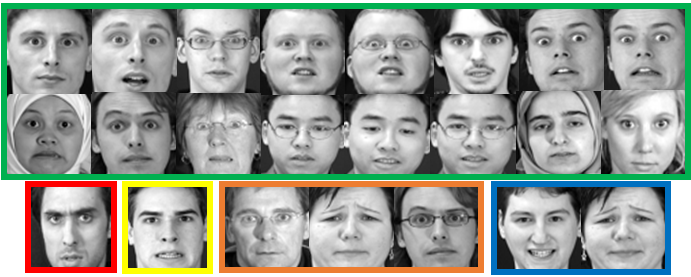}
\caption{\textbf{All Failure Cases with Fear in the MMI database.} Our deep network predicted fear to surprise (green box), anger (red box), disgust (yellow box), sadness (orange box), and happiness (blue box).}
\label{fig:fail}
\end{center}
\vspace{-4mm}
\end{figure}

We compared two networks and the combined model in Figure \ref{fig:mmi_compare}. The two networks were complementary to each other for most of the emotions. However, with fear, our algorithm was degraded. This is also shown in the confusion matrix in Table \ref{tb:mmi_cm}. We observed that the accuracy for fear was much lower than other emotions. In particular, most of the fear emotions were confused with surprise. To determine why this phenomenon appears, we checked all the failure cases as shown in Figure \ref{fig:fail}.

The results indicated that a variety of facial expressions represented fear. Many cases were similar to surprise or sadness. To train for these various expressions, various kinds of training data are additionally required. However, we had only 27 subjects for training data. (Three subjects were used for validation.) Unfortunately, performance of deep learning techniques highly depends on the quality of training data, so our accuracy with fear was not good enough.

\section{Conclusion}
We presented two deep network models that collaborate with each other. The first network was DTAN, which was based on multiple frames, while the second network was DTGN, which extracted useful temporal geometric features from raw facial landmark points. We showed that the filters learned by the DTAN in the first layer have the ability to obtain the difference between the input frames. Furthermore, the important landmark points extracted by DTGN were also shown. As a result, we achieved best recognition rates using the integrated deep network with the CK+ and Oulu-CASIA databases. However, in the MMI database, our algorithm had a lower accuracy because there were only 30 subjects, which is too small a sample size to make correct predictions using deep learning models. 


{\small
\bibliographystyle{ieee}
\bibliography{egpaper_for_review}
}

\end{document}